\documentclass[letterpaper]{article} 
\usepackage{aaai25}  
\usepackage{times}  
\usepackage{helvet}  
\usepackage{courier}  
\usepackage[hyphens]{url}  
\usepackage{graphicx} 
\usepackage{natbib}  
\usepackage{caption} 
\usepackage{algorithm}
\usepackage{algorithmic}
\usepackage{amsfonts}
\usepackage{multirow} 
\usepackage{booktabs}

\usepackage{newfloat}
\usepackage{listings}
\DeclareCaptionStyle{ruled}{labelfont=normalfont,labelsep=colon,strut=off} 
\floatstyle{ruled}
\newfloat{listing}{tb}{lst}{}
\floatname{listing}{Listing}
\nocopyright

\title{PSPO*: An Effective Process-supervised Policy Optimization for Reasoning Alignment}

\author{Jiawei Li, Xinyue Liang, Junlong Zhang, Yizhe Yang, Chong Feng, Yang Gao\thanks{Corresponding author.}\\
        School of Computer Science and Technology,\\
        Beijing Institute of Technology, Beijing, China \\
        \texttt{\{jwli, xyliang, takizhang, yizheyang, fengchong, gyang\}@bit.edu.cn}}

\usepackage{bibentry}

\begin{document}

\maketitle

\begin{abstract}
Process supervision enhances the performance of large language models in reasoning tasks by providing feedback at each step of chain-of-thought reasoning. However, due to the lack of effective process supervision methods, even advanced large language models are prone to logical errors and redundant reasoning.
We claim that the effectiveness of process supervision significantly depends on both the accuracy and the length of reasoning chains. Moreover, we identify that these factors exhibit a nonlinear relationship with the overall reward score of the reasoning process. 
Inspired by these insights, we propose a novel process supervision paradigm, PSPO*, which systematically outlines the workflow from reward model training to policy optimization, and highlights the importance of nonlinear rewards in process supervision. Based on PSPO*, we develop the PSPO-WRS, which considers the number of reasoning steps in determining reward scores and utilizes an adjusted Weibull distribution for nonlinear reward shaping. Experimental results on six mathematical reasoning datasets demonstrate that PSPO-WRS consistently outperforms current mainstream models. Our code can be found at \url{https://github.com/DIRECT-BIT/PSPO}\footnote{This paper is submitted to AAAI-2025.}.
\end{abstract}

%

\section{Introduction}

Solving tasks that require complex reasoning, such as mathematical problems, remains a significant challenge for large language models (LLMs). Chain-of-Thought (CoT) has shown potential in significantly improving the performance of models on complex reasoning tasks by guiding them to break down and solve problems step by step~\cite{DBLP:conf/nips/Wei0SBIXCLZ22}. Studies have shown that an effective reasoning chain can significantly improve a model's performance on downstream tasks. Conversely, unreliable reasoning chains can mislead the model and produce incorrect results~\cite{DBLP:conf/iclr/0002WSLCNCZ23,DBLP:journals/corr/abs-2401-04925}. 
Therefore, quantifying an accurate reasoning process is crucial for effectively addressing the complex reasoning task.
Using process supervision to achieve reasoning alignment is considered an effective way to quantifying reasoning process~\cite{DBLP:journals/corr/abs-2305-20050,DBLP:journals/corr/abs-2308-09583,liang-etal-2024-bit}. In process supervision, each step receives precise supervision. It provides feedback for each individual step in a chain-of-thought and rewards the model for aligning with the human-approved reasoning process~\cite{DBLP:journals/corr/abs-2211-14275,DBLP:journals/corr/abs-2305-20050}.

\begin{figure}[t]
\centerline{\includegraphics[width=\linewidth]{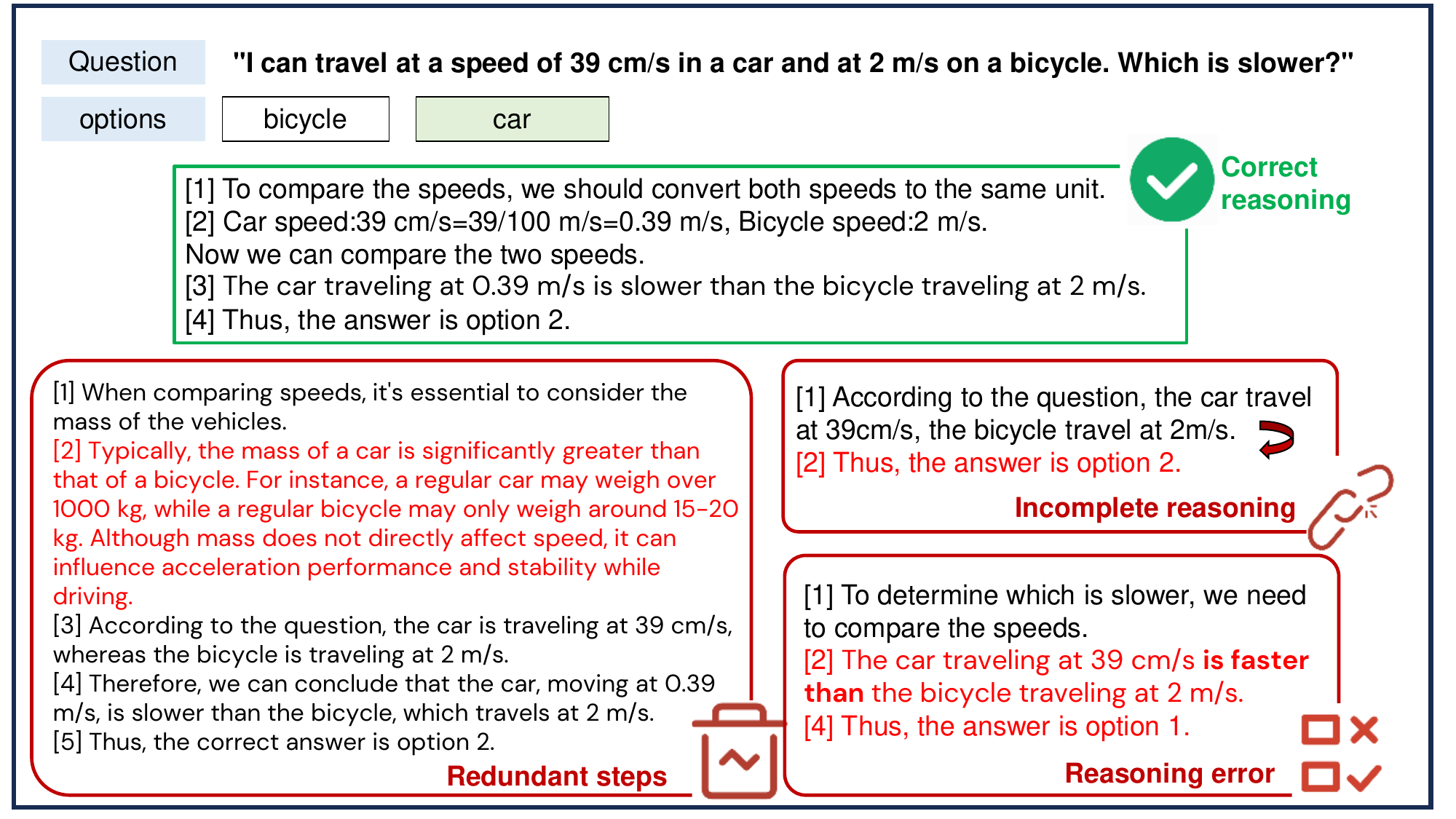}} 
\caption{An example from QQA dataset. 
The reasoning error solution has an error in step[2] where the model confuses the concept of time period and time point, resulting in a wrong answer. The imcomplete reasoning solution simply jump to the final answer after summarizing the problem, which is incomplete and unreasonable. And the redundant steps generate too much noise.}
\label{intro-example}
\end{figure}

The implementation of process supervision involves training process-supervised reward models (PRMs) and computing a single PRM score by aggregating scores from multiple reasoning chains~\cite{DBLP:journals/corr/abs-2211-14275}. ~\citet{DBLP:journals/corr/abs-2305-20050} employed manual annotation to obtain 800K aligned data for training the PRMs. They defined the PRM score for a reasoning chain as the probability that every step is correct under the PRM, which is calculated as the product of the probabilities of correctness for each step. ~\citet{DBLP:journals/corr/abs-2312-08935} avoided dividing the reasoning chain into multiple steps for PRM training; instead, they trained the PRM to evaluate the entire reasoning chain collectively. However, these methods overlook the impact of the length of the reasoning chain on the PRM score. 
In this paper, we propose that the effectiveness of process supervision is influenced by both the accuracy and the length of reasoning chains. As illustrated in Figure~\ref{intro-example},  inaccurate reasoning chains those that are excessively long or short can lead to wrong outcomes. Consequently, we aim to develop more effective process supervision methods to address this issue.

In this work, we propose a process-supervised policy optimization (PSPO*) paradigm. PSPO* systematically standardizes the overall process supervision workflow, including reward model training and policy optimization. During reward model training phase, we compare the differences between the outcome-supervised reward model (ORMs) based on the Bradley-Terry model and the PRMs trained using multi-class classification. Subsequently, we standardize the training procedure for the PRMs. 
During policy optimization phase, propose employing an accumulation function that is correlated with the accuracy and the length of reasoning chains to calculate their reward scores. This methods aims to achieve more efficient process supervision.

Additionally, we are the first to assert that the reward score in the reasoning alignment is nonlinear. To enhance this nonlinear characteristic, we propose employing a nonlinear accumulation function to calculate process-supervised reward scores. Furthermore, we propose employing nonlinear reward shaping to effectively incorporate prior knowledge from process supervision into the rewards. 

To validate the effectiveness of the aforementioned methods, we concrete the PSPO* paradigm using adjustable Weibull distribution reward shaping (WRS) named PSPO-WRS. In the PSPO-WRS, to comprehensively account for the impact of both the accuracy and the number of inference steps on the overall reward, we compute the overall reward score of the reasoning process by multiplying the reward scores of individual reasoning steps and then normalize them with respect to the steps. Additionally, we employ prior knowledge from process supervision to construct an adjustable Weibull distribution, which is incorporated into the reward shaping to enhance the nonlinear characteristics of the reward scores. Experimental results demonstrate that PSPO-WRS exhibits superior performance across various datasets, thereby affirming the effectiveness of the PSPO* method. The experimental analysis validates the hypothesis that the reward score in the reasoning alignment is nonlinear.

Our main contributions are as follows:
\begin{itemize}
\item We propose PSPO*, a novel paradigm for process supervision to achieve reasoning alignment. PSPO* encompasses the entire workflow for process supervision from reward model training to policy optimization.
\item We are the first to assert that the reward score in the reasoning alignment is nonlinear. We propose using nonlinear accumulation functions and incorporating prior supervision knowledge through nonlinear reward shaping to enhance this characteristic.
\item Building upon PSPO*, we propose the PSPO-WRS method, which utilizes a nonlinear reward accumulation function correlated with the accuracy and quantity of reasoning steps, and employs adjusted Weibull distribution for nonlinear reward shaping. Experimental results demonstrate the effectiveness of PSPO-WRS and further validate the correctness of the PSPO* framework.
\end{itemize}

\section{Related Works}
\subsection{Alignment Methods}
LLMs trained on large-scale corpora have demonstrated outstanding reasoning abilities. Despite their significant performance, these models are prone to limitations like misunderstandings of human instructions, logical errors that do not conform to common sense, and providing inaccurate information. Therefore, aligning LLMs with human expectations has become a focus of research \cite{DBLP:journals/corr/abs-2307-12966}. The classical Reinforcement Learning from Human Feedback (RLHF) framework for alignment \cite{DBLP:conf/nips/Ouyang0JAWMZASR22} typically consists of two stages: (1) Reward training based on human feedback, where a reward function is learned. (ii) Optimization, where the policy model uses a reinforcement learning algorithm(Proximal Policy Optimization, PPO \cite{DBLP:journals/corr/SchulmanWDRK17}) to optimize the rewards learned in the previous step. However, RLHF is a complex and often unstable procedure, \citet{DBLP:conf/nips/RafailovSMMEF23} introduced Direct Preference Optimization (DPO), which parameterizes reward model, solving the standard RLHF problem with a simple classification loss. In the classical RLHF framework, PPO is employed to learn from sparse, sentence-level rewards, to optimize its open-source implementation, \citet{DBLP:journals/corr/abs-2404-18922} introduced a framework named Reinforced Token Optimization (RTO), which learns a token-wise reward function from preference data and performs policy optimization based on the learned reward signal. \citet{DBLP:conf/aistats/AzarGPMRVC24} further proposed a general objective, termed $\Psi$PO, and revealed that in principle RLHF and DPO can be both prone to overfitting as both methods rely on the strong assumption that pairwise preferences can be substituted with ELo-score (pointwise rewards) via a Bradley-Terry modelization. 
Therefore, we propose a novel PSPO* paradigm for process supervision involving step-level pointwise rewards and policy optimization.

\subsection{Reasoning Process Alignment}
LLMs have significantly progressed in complex multi-step reasoning tasks in recent years. Studies have shown that step-wise reasoning, such as CoT and ToT, can improve model performance on reasoning tasks \citep{DBLP:journals/corr/abs-2110-14168,wei2023chainofthought,yao2024tree} as it helps to decompose complex problems and guide models toward solutions. Such methods can enhance the reasoning abilities of LLMs, particularly in mathematical reasoning, where accuracy requires a precise chain of thought \citep{kojima2023largelanguagemodelszeroshot}. \citet{uesato2022solvingmathwordproblems} and \citet{DBLP:journals/corr/abs-2305-20050} introduced both the Outcome-supervised Reward Model (which provides feedback for the final result) and the Process-supervised Reward Model (which provides feedback for step-level reasoning process), and demonstrated that process-based supervision is necessary for correct reasoning steps and avoiding false positives solutions that reach correct answer with incorrect reasoning. As ensuring the correctness of each reasoning step is critical, \citet{DBLP:journals/corr/abs-2406-18629} proposed Step-DPO, which treats individual reasoning steps as units for preference optimization rather than evaluating answers holistically. However, these methods require a large amount of supervised data, which is very costly. \citet{DBLP:journals/corr/abs-2406-03816} further proposed a reinforced self-training approach, called ReST-MCTS*, based on integrating process reward guidance with tree search MCTS* for collecting higher-quality reasoning traces as well as per-step value to train policy and reward models.
Despite their success, the potential for facilitating the model's ability to identify more critical behaviors by adjusting the reward structure remains unexplored. To fill this gap, we implement a nonlinear reward shaping function upon PSPO*, terms PSPO-WRS method, and employ adjusted Weibull distribution to validate this framework.

\section{The Paradigm for Process Supervision}
Similar to the standard RLHF paradigm, the process supervision based on human feedback primarily consists of two stages: learning the reward model and optimizing the policy based on the learned reward model~\cite{DBLP:conf/aistats/AzarGPMRVC24}. Considering the characteristics of process supervision, in this section, we will provide a detailed introduction to these two stages and propose a general theoretical paradigm applicable to process supervision.

\subsection{Reward Model for Process Supervision}
In the process of training the outcome-supervised reward models (ORMs), annotators are required to distinguish between human-preferred and non-preferred responses in the candidate responses for a given input~\cite{DBLP:conf/nips/Ouyang0JAWMZASR22}. Based on this annotated data, researchers typically employ the Bradley-Terry model to construct a classification model, and subsequently train the reward model using pairwise loss~\cite{DBLP:journals/corr/abs-2307-12966,DBLP:conf/nips/RafailovSMMEF23}. For a given context $x$ and action $y$, the Bradley-Terry model represents the preference function $p(y_w \succ y_l)$ as a sigmoid of the difference of rewards:
$$p(y_w \succ y_l \mid x)=\sigma(r(x,y_w)-r(x,y_l)),$$
where $\sigma(\cdot)$ denotes the sigmoid function and plays the role of normalisation, $r(x,y)$ denotes the pointwise reward of $y$ given $x$, $y_w$ denotes the human-preferred responses and $y_l$ denotes the non-preferred responses. Given the dataset $\mathcal{D}=(x_i,y_{w,i} \succ y_{l,i})_{i=1}^N$ one can learn the reward function by optimizing the following logistic regression loss:
$$\mathcal L(x)=-\mathbb{E}_{(x,y_w,y_l) \sim \mathcal{D}}[log(p(y_w \succ y_l \mid x))].$$

\begin{figure}[t]
\centering
\includegraphics[width=\columnwidth]{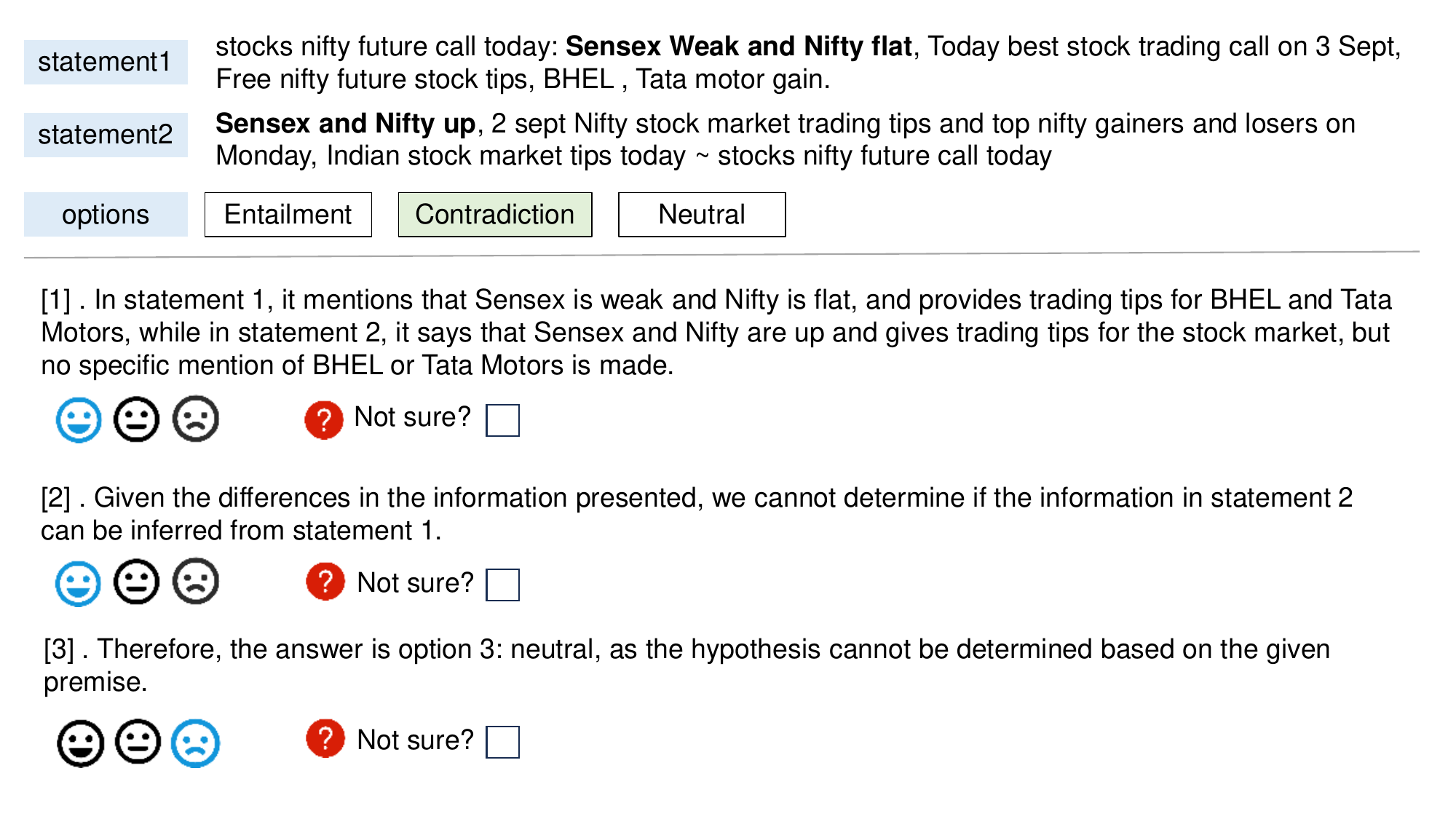} 
\caption{The data annotation approach for PRM. Unlike ORM, the annotation approach of PRM cannot generate pairwise preference data, thus precluding the use of the Bradley-Terry method for training the reward model.}
\label{fig:annotators}
\end{figure}

In the process of training the PRMs, annotators are required to assess the correctness of each step in the model-generated solutions. Specifically, as illustrated in Figure~\ref{fig:annotators}, annotators typically need to determine whether the current reasoning step is negative, neutral, or positive, and correspondingly select from [-1, 0, 1]~\cite{DBLP:journals/corr/abs-2305-20050,DBLP:journals/corr/abs-2310-10080}. These annotated data are subsequently used to train the reward model, thereby enhancing its capability to distinguish and classify negative, neutral, and positive steps. However, due to the absence of pairwise comparison data regarding human preferences in this process, the Bradley-Terry model cannot be employed to construct a classification model. Here we redefine the training process of PRM.

In light of the coherence of reasoning steps, evaluating the accuracy of the $k$-th reasoning step $y^k$ necessitates the simultaneous consideration of the input $x$ and the preceding $k$ reasoning steps $y_{pre}^k$ as context. The reward model maps these inputs to an $n$-dimensional vector $z$, which encompasses the scores or raw outputs for each category. Formally, this can be represented as:
\begin{equation}\label{eq1}
    z = r(x,y_{pre}^{k},y^k;\theta),
\end{equation}
where $\theta$ denotes the parameters of the reward model, and $r(\cdot)$ denotes the reward model. We employ an activation function to transform the model outputs to a probability distribution:
\begin{equation}\label{eq2}
    p(z_i) = \sigma(z_i),
\end{equation}
where $p(z_i)$ denotes represents the probability that the current step belongs to category $i$, and $\sigma( \cdot )$ denotes the activation function, which is typically the softmax function:
\begin{equation}\label{eq3}
    softmax(z_i)=\frac{e^{z_i}}{\sum_{j=1}^ne^{z_j}},
\end{equation}
where $n$ denotes the total number of categories.
The training of reward models typically employs the cross-entropy loss function to quantify the divergence between the predicted probability distribution and the true labels. Let $\mathbf{z}$ denote the one-hot encoded vector of the true labels. The training process is then formulated as follows:
\begin{equation}\label{eq4}
    \mathcal{L(\theta)}=-\mathbb{E}_{(x,y) \sim \mathcal{D}}[\sum_{i=1}^N\mathbf{z}log(p({z_i}))].
\end{equation}
Ultimately, the reward model $r(x,y_{pre}^k,y^k)$ predicts the probabilities of the current reasoning step $y$ belonging to various categories. The probability assigned to the positive category is then used as the reward score for the current reasoning step, as follows:
\begin{equation}\label{eq5}
    R_{s}^k=p(z_{i=1}),
\end{equation}
where $R_{s}^k$ denotes the reward score of the $k$-th reasoning step.

\begin{figure*}[h]
\centering
\includegraphics[width=0.95\linewidth]{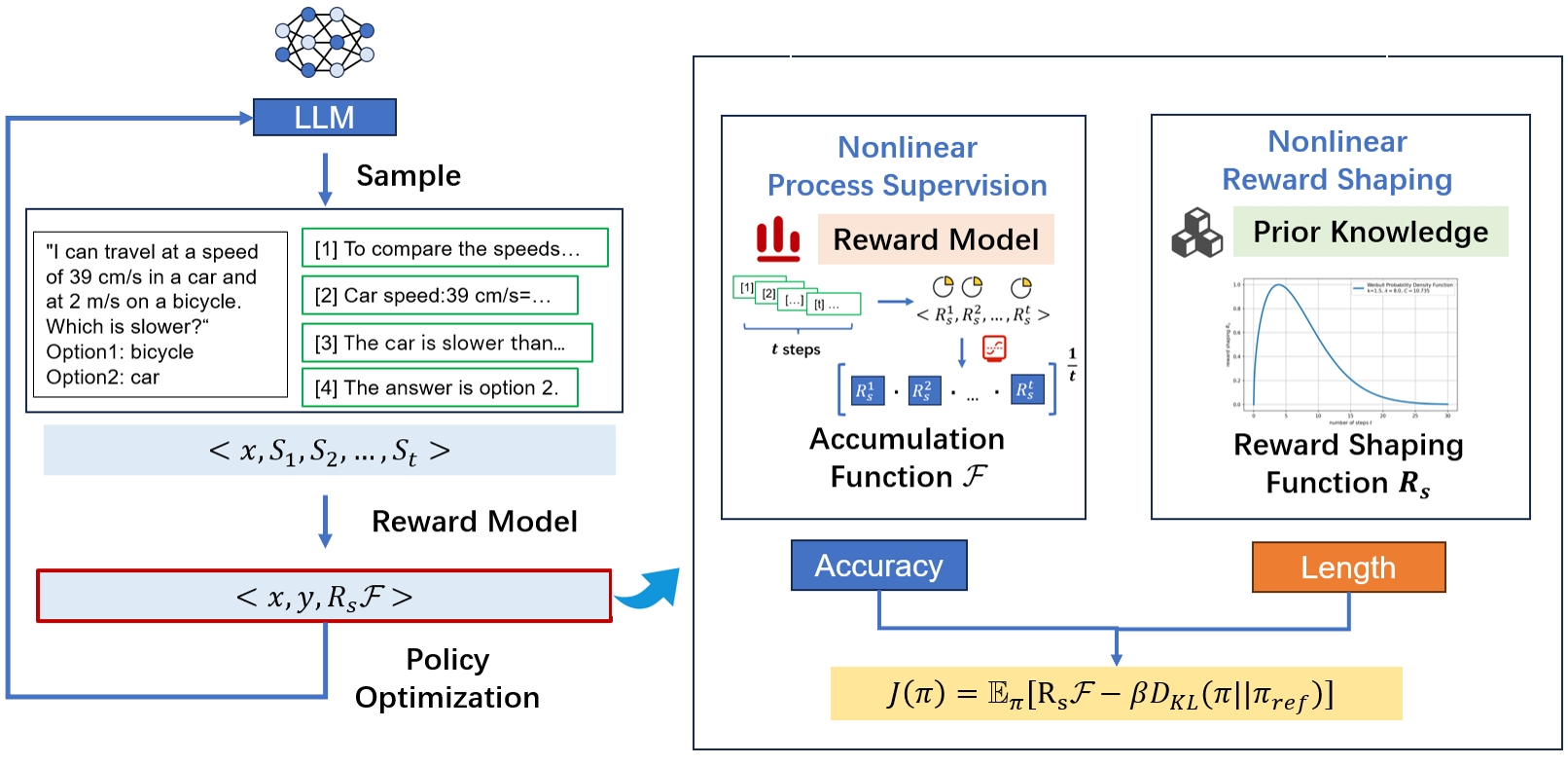} 
\caption{The overall method of PSPO*. In the workflow of process supervision, we encompass a nonlinear accumulation function that is correlated with the accuracy of the reasoning chains and nonlinear reward shaping that shapes the rewards for the length of the reasoning chains. }
\label{fig:main}
\end{figure*}

\subsection{Policy Optimization with the Reward Score}
Based on the reward model $r(x,y_{pre}^k,y^k)$, the score $R_s^k$ for the current $k$-th reasoning step only reflects the quality of that individual step and not the whole reasoning process. In process supervision, the reward for the whole reasoning process can only be evaluated by accumulating the scores of all reasoning steps. 
We define the overall reward score for the whole reasoning process as $R(x,y)$, and let $\mathcal{F}$ be the accumulation function. In previous studies, the construction of the reward function (F) typically only considered the impact of accuracy for the overall reward score $R(x,y)$ in the reasoning chains~\cite{DBLP:journals/corr/abs-2305-20050}. Our contribution lies in proposing that, the accumulation function $\mathcal{F}$ should simultaneously account for both the accuracy and the length of reasoning chains in process supervision. Specifically, we define the length of reasoning chains by the number of steps in the reasoning process, then:
\begin{equation}\label{eq6}
    R(x,y)=\mathcal{F}(R_s^1,R_s^2,\cdots,R_s^t),
\end{equation}
where $t$ denotes the total number of reasoning steps.

The objective of process supervision is to optimize the policy function $\pi \in \Delta(x,y)$  through the overall reward score $R(x,y)$ of the reasoning process, thereby maximizing the expected reward. Simultaneously, it aims to minimize the KL divergence between $\pi$ and the reference policy $\pi_{ref} \in \Delta(x,y)$:
\begin{equation}\label{eq7}
    J(\pi)=\mathbb{E}_\pi[R(x,y)-\beta D_{KL}(\pi \parallel \pi_{ref} )],
\end{equation}
where $\beta$ is a hyperparameter used to limit the difference between the new and reference policies, balancing the exploration and exploitation of the policy.

\subsection{Nonlinear Reward Shaping}
In traditional RLHF processes, there is typically a linear trend between the quality of the model-generated response and the reward score, such that higher response quality corresponds to higher reward scores. In process supervision, researchers commonly construct the accumulation function $\mathcal{F}$ for the reasoning process based on this linear trend. For instance, \citet{DBLP:journals/corr/abs-2305-20050} proposed that using the product of the reward scores for each reasoning step as the accumulation function $\mathcal{F}$, thereby modeling the overall reward score for the entire reasoning process, as follows:
$$ R(x,y)=\prod_{j=1}^tP(y^j=1|x^j, y_{pre}^j).$$
However, when the number of reasoning steps is not fixed, the overall reward score is influenced by the number of reasoning steps. As the correctness probability is decimal, the more steps involved in reasoning, the smaller the product of probabilities, resulting in lower rewards, which leads to a tendency for the policy to subsequently generate fewer reasoning steps.

From Equation~\ref{eq6}, it can be observed that the accumulation function $\mathcal{F}$ is not only related to the quality of the reasoning steps but also to the length of the reasoning chains. The accumulation function $\mathcal{F}$ based solely on the linear trend of reasoning step quality is insufficient for comprehensively modeling process supervision. Therefore, further refinement of the accumulation function is necessary.

A key contribution of our work is the introduction of nonlinear reward shaping to refine the accumulation function. Reward shaping refers to the process of transforming prior knowledge into additional rewards, guiding the policy to learn faster and better by combining the original rewards with these new rewards~\cite{DBLP:conf/nips/HuWJWCH0F20}. In process supervision, to enable the policy to better distinguish critical behaviors, we propose to apply \textit{nonlinear reward shaping}.  Nonlinear functions can provide larger rewards for behaviors with particularly significant outcomes and smaller rewards for routine or minor behaviors, thereby amplifying or diminishing the impact of certain rewards~\cite{DBLP:conf/nips/LevinePK11}. The final policy optimization is:
\begin{equation}\label{eq8}
    J(\pi)=\mathbb{E}_\pi[R_sR(x,y)-\beta D_{KL}(\pi \parallel \pi_{ref} )],
\end{equation}
where $R_s$ is a nonlinear function used for reward shaping. Specifically, the method proposed by~\citet{DBLP:journals/corr/abs-2305-20050} can be viewed as a simplified version of the PSPO* paradigm. In their method, the value for reward shaping is specifically set to 1, and the construction of the accumulation function does not take into account the impact of the length of reasoning chains on the reward score.

In the next section, we utilize the prior knowledge from process supervision to perform nonlinear reward shaping using the adjusted Weibull distribution, demonstrating the validity of this view.

\section{PSPO-WRS: Process-supervised Policy Optimization with Nonlinear Reward Shaping
}
\label{sec:weibull}
In process supervision, there is a nonlinear relationship between the number of reasoning steps and the overall reward score. The goal of the CoT reasoning is typically to solve the problem while minimizing computational complexity~\cite{DBLP:conf/nips/Wei0SBIXCLZ22}. Fewer reasoning steps imply higher efficiency, but this does not always correlate with higher accuracy or correctness. Conversely, a reasoning process with more steps might achieve greater accuracy, but at the cost of lower efficiency. Based on this prior knowledge, we employ the Adjusted Weibull distribution to shape the rewards for the number of reasoning steps. The reward shaping function is as follows:
\begin{equation}\label{eq9}
    R_s=C*\frac{k}{\lambda}(\frac{t}{\lambda})^{k-1}e^{-(t/\lambda)^k},
\end{equation}
where $C$ is a constant coefficient used to adjust the overall reward score, $\lambda$ is the scale parameter, which determines the spread of the distribution, and $k$ is the shape parameter, which dictates the shape of the distribution.

Additionally, for the accumulation function $\mathcal{F}$, to eliminate the linear trend and account for the number of reasoning steps, we standardized the step count based on the method proposed by~\citet{DBLP:journals/corr/abs-2305-20050}, specifically:
\begin{equation}\label{eq10}
    \mathcal{F} = [\prod_{j=1}^tP(y^j=1|x^j, y_{pre}^j)]^{1/t}.
\end{equation}

Finally, we propose process supervision based on adjusted Weibull Reward Shaping (PSPO-WRS):
\begin{equation}\label{eq11}
    J(\pi)=\mathbb{E}_\pi[R_s\mathcal{F}-\beta D_{KL}(\pi \parallel \pi_{ref} )].
\end{equation}

The PSPO-WRS introduces nonlinear reward shaping, integrating both the accuracy and the length of reasoning chains into process supervision. In the experimental section, we will demonstrate the effectiveness of PSPO-WRS.

\section{Experimental Results}
\label{experiment}

\begin{table}[]
\centering
\begin{tabular}{p{1.5cm}p{1cm}p{0.6cm}p{0.6cm}p{0.6cm}l}
\toprule
     \multirow{2}{*}{Datasets}& \multirow{2}{*}{Cases} & \multicolumn{4}{c}{Human labeled}                                        \\
     &       & Pos. & Neu. & Neg. & \begin{tabular}[c]{@{}l@{}}Steps\end{tabular} \\
\midrule
AwpNLI  & 1622 & 4334   & 822   & 1669   & 7109  \\
NewsNLI & 1643  & 3358    &910     & 2870    &7502  \\
RedditNLI &1152   &3074     &507     &958   &4674  \\
RTE\_Quant &1324   &3363     & 290    & 914    &4817  \\
StressTest &1369   &2598     &723     &1921     &5696  \\
QQA  &1394   &3937     &184     &1778     &6424     \\                          \bottomrule
\end{tabular}
\caption{The step data labeled by human. "Cases" is the number of solutions generated by models, "Pos.", "Neu.", and "Neg." are the number of positive, neutral, and negative labels after labeling, respectively, "Steps" is the total number of reasoning steps taken to solve all the questions in the dataset.}
\label{tab:data_table}
\end{table}

\begin{table*}[t]
\centering
\begin{tabular}{lcccccc}
\toprule
Models    & AwpNLI           & NewsNLI          & RedditNLI        & RTE-Quant        & StressTest       & QQA              \\ \midrule
Llama2-7B~\cite{DBLP:journals/corr/abs-2307-09288} & 1.47\%           & 0.47\%           & 0.40\%           & 0.86\%           & 1.36\%           & 3.70\%           \\
BLOOMZ~\cite{DBLP:conf/acl/MuennighoffWSRB23}    & 48.04\%          & 54.46\%          & 37.20\%          & 47.64\%          & 31.22\%          & 51.85\%          \\
Abel-7B~\cite{abel}   & 55.82\%          & 50.75\%          & 47.20\%          & 56.67\%          & 30.87\%          & 48.14\%          \\
Llama3.1-8B~\cite{dubey2024llama3herdmodels} & 66.18\%          & 62.91\%          & 39.60\%          & 48.93\% & 13.04\% & 50.62\% \\
Qwen1.5-7B-chat~\cite{yang2024qwen2technicalreport} & 54.90\%          & 54.93\%          & 40.00\%          & 21.13\% & 27.32\% & 46.30\% \\
CN-PPO~\cite{liang-etal-2024-bit}    & 82.35\%          & 61.97\% & 63.20\%          & 63.52\%          & 46.30\%          & 48.77\%          \\
\midrule
PSPO-WRS (Ours) & \textbf{86.76\%} &  \textbf{64.91\%}         & \textbf{67.60\%} & \textbf{71.57\%}          & \textbf{52.29\%}         & \textbf{54.70\%}          \\ \bottomrule
\end{tabular}
\caption{Performance of baseline models. The prompt of GPT-3.5* has added explanations for options such as "entailment" compared to GPT-3.5.}
\label{tab:main_results}
\end{table*}

\subsection{Human-data Collection for Training Reward Model}
Training a robust reward model requires a balanced label distribution. While the steps generated by GPT-3.5 predominantly feature positive labels, we included additional reasoning step candidates from other LLMs, such as Abel-7b, to provide more negative examples and achieve label balance.  Human labelers would evaluate the given steps by their correctness, and correct answers to the question are provided as a reference. The statistics of datasets are shown in Table~\ref{tab:data_table}.

\paragraph{Step Labelling Criteria} 
Each reasoning step is evaluated and assigned a label based on its correctness: `positive' (score of `1'), `neutral' (score of `0'), and `negative' (score of `-1'). A step receives a positive score if it accurately meets logical and computational requirements, correctly interprets the task, and contributes to deriving the correct answer. A neutral score is awarded if the step is correct but does not aid in reaching the correct conclusion. Conversely, steps that contain logical, computational, or factual inaccuracies, or are irrelevant to the given context and question, are assigned a negative score of -1.

\subsection{Experimental Setups}

\paragraph{Datasets} 
We adopt the MATH dataset, which includes AwpNLI, NewsNLI, RedditNLI, RTE-Quant, StressTest, and QQA datasets as reported by~\citet{chen-etal-2023-improving}. These datasets are further expanded using the GPT-3.5 API, as detailed in Table~\ref{tab:data_table}. The training dataset for the reward model is primarily composed of data labeled as `1'. To ensure a balanced dataset, steps labeled `0' and `-1' are replicated 2-3 times, yielding a final count of 16,587 positive, 11,072 neutral, and 16,236 negative steps. For evaluation, 20\% of the dataset is designated as test sets.

\begin{figure}[t]
\centering
\includegraphics[width=\columnwidth]{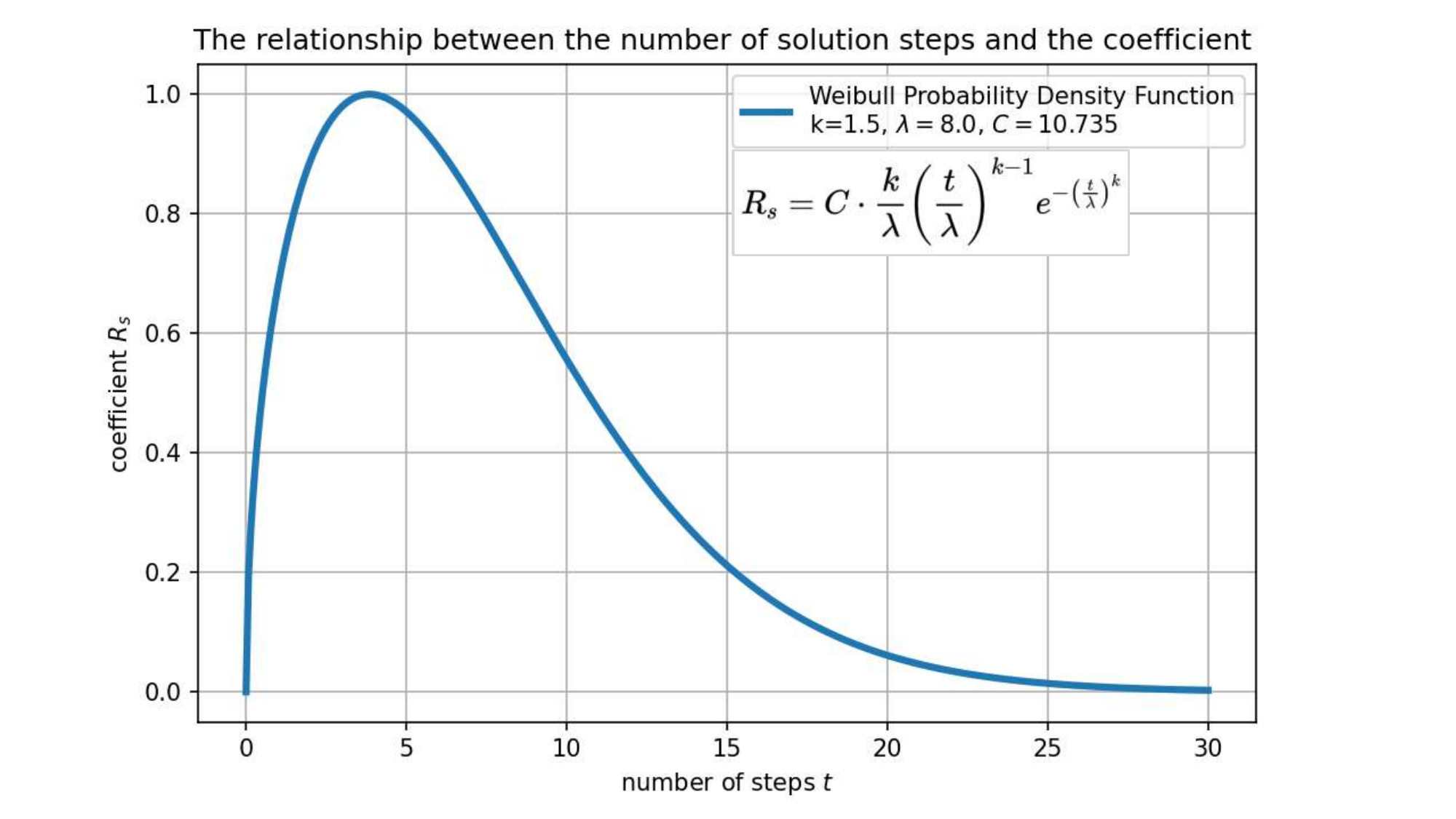} 
\caption{The adjusted weibull distribution. Prameter settings are: $C=10.735$, $k=1.5$, and $\lambda=8.0$.}
\label{fig:distribution}
\end{figure}

\paragraph{Metrics and Parameters setting} 
The evaluation metric utilized is the average micro-F1 score on the test dataset because it balances precision and recall, providing a more comprehensive measure of model performance. We employ Abel-7B as the baseline model, which has been fine-tuned on a substantial portion of the MATH dataset for generating chain-of-thought reasoning in mathematical problem-solving~\cite{abel}. The reward model is trained on the BERT-large~\cite{DBLP:conf/naacl/DevlinCLT19} due to its proven efficacy in classification tasks~\citep{DBLP:conf/icml/GaoSH23}. We trained the reward model over 10 epochs with a learning rate of 2e-5, a warmup rate of 0.05, and a maximum sequence length of 256.PPO training uses Lora~\cite{DBLP:conf/iclr/HuSWALWWC22} with a learning rate of 1.41e-5 and a maximum of 512 tokens. On a dataset of 5470 entries, each epoch averages 55 hours on four NVIDIA A100 GPUs. In our PSPO-WRS method, the parameters are set as follows: $C=10.735$, $k=1.5$, and $\lambda=8.0$. The function distribution is illustrated in Figure~\ref{fig:distribution}.

\subsection{Overall Results}

\paragraph{Main Results} Table~\ref{tab:main_results} compares the performance of our method with that of current mainstream LLMs on the mathematic reasoning processing tasks. In comparisons with mainstream models of equivalent size, PSPO-WRS achieves optimal performance in the AwpNLI, NewsNLI, RedditNLI, RTE\_Quant, StressTest, and QQA tasks. 
Specifically, compared to our baseline model Abel-7B, our model achieved significant accuracy improvements of 30.94\%, 14.16\%, 20.4\%, 14.9\%, 21.42\%, and 6.56\% across six tasks. These results validate the effectiveness of the methods proposed in this study.

\begin{figure}[t]
\centering
\includegraphics[width=\columnwidth]{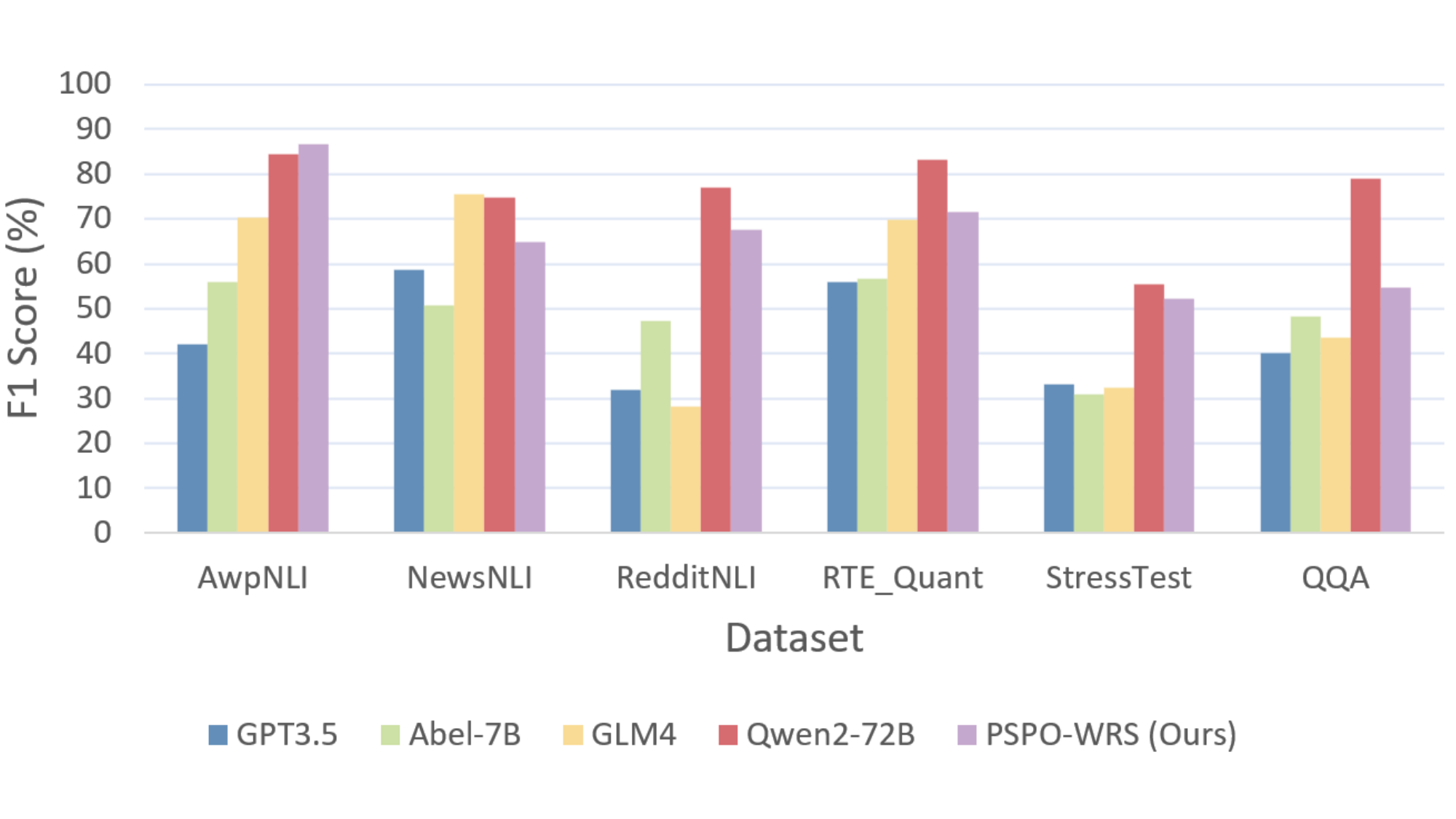} 
\caption{The results compared with ultra LLMs. It is noteworthy that our model outperforms ultra LLMs in most scenarios with only 7B parameters.}
\label{fig:big_models}
\end{figure}

\paragraph{PSPO-WRS exhibits exceptional performance even when compared with ultra LLMs.} We conducted a comparative analysis of PSPO-WRS against mainstream ultra LLMs, as detailed in Figure~\ref{fig:big_models}. Across all evaluated datasets, the PSPO-WRS significantly outperformed GPT-3.5~\cite{DBLP:conf/nips/Ouyang0JAWMZASR22}. Relative to GLM4~\cite{DBLP:journals/corr/abs-2406-12793}, our model showed slightly weaker performance on the NewsNLI dataset, yet exhibited superior performance on other datasets. Against the more robust reasoning capabilities of Qwen2-72B~\cite{yang2024qwen2technicalreport}, PSPO-WRS also showed its strengths in the AWPNLI dataset and demonstrated comparable performance on additional datasets. Notably, although the baseline model Abel-7B~\cite{abel} of PSPO-WRS is far outperformed by these ultra LLMs in terms of raw performance, our process supervision method effectively bridges this gap, showcasing its efficacy.

\begin{table}[t]
\centering
\begin{tabular}{ l | cccc }
\hline
Dataset  & w/o nonlinear  & PSPO-WRS \\
\hline
anli-r1   & 38.30\%  & \textbf{43.60\%} \\
anli-r2   & 32.33\%  & \textbf{40.80\%} \\
anli-r3  & 34.70\%  & \textbf{40.33\%} \\
glue-mini-matched   & 42.20\%  & \textbf{55.60\%} \\
glue-wnli  & 47.89\%  & \textbf{49.30\%} \\
stress-negation & 34.20\%  & \textbf{45.53\%} \\
\hline
\end{tabular}
\caption{
Out-of-distribution generalization of PSPO-WRS. The experimental results demonstrate that the success of PSPO* is not due to fitting to the evaluation datasets, but rather to a genuine enhancement of the model's mathematical reasoning capabilities.
}
\label{tab:ood}
\end{table}

\paragraph{The performance improvement of PSPO-WRS is NOT due to overfitting.} We conducted experiments on test sets outside the distribution of the training data for the reward model to validate that the performance gains of PSPO-WRS are not due to overfitting. As shown in Table~\ref{tab:ood}, PSPO-WRS consistently outperforms the baseline model across six datasets not included in the training distribution. The results demonstrate that PSPO-WRS also exhibits strong reasoning capabilities on out-of-distribution mathematical datasets.

\begin{table}[t]
\centering
\begin{tabular}{ l | cccc }
\toprule
Dataset  & w/o nonlinear  & PSPO-WRS \\
\midrule
AwpNLI   & 69.12\%  & \textbf{64.91\%} \\
 NewsNLI   & 56.81\%  & \textbf{67.60\%} \\
 RedditNLI  & 56.40\%  & \textbf{71.57\%} \\
 RTE\_Quant   & 59.66\%  & \textbf{52.29\%} \\
 StressTest  & 41.34\%  & \textbf{53.57\%} \\
QQA & 54.94\%  & \textbf{54.70\%} \\
\midrule
Num. of steps & 2.624 & 3.051 \\
\bottomrule
\end{tabular}
\caption{
Comparison results indicate that our proposed RLHF outperforms all datasets and can generate more completed reasoning steps. 
}
\label{tab:ablation}
\end{table}

\subsection{Ablation Analysis}
\paragraph{Process supervision is nonlinear.} In the ablation study, we evaluated the effectiveness of nonlinear rewards within the PSPO-WRS. Notably, upon the removal of the nonlinear reward, our method degenerated into the process-supervised reward method proposed by~\citet{DBLP:journals/corr/abs-2305-20050}, and the results are illustrated in Table~\ref{tab:ablation}. Following the removal of the nonlinearity module, there is a noticeable decline in the performance of PSPO-WRS. The experimental results indicate that the introduction of nonlinear rewards led to significant performance improvements across all evaluated datasets. These findings underscore the critical importance of nonlinear rewards in the process supervision.

\begin{figure}[t]
\centering
\begin{minipage}[t]{\linewidth}
    \centering
    \includegraphics[width=0.95\textwidth]{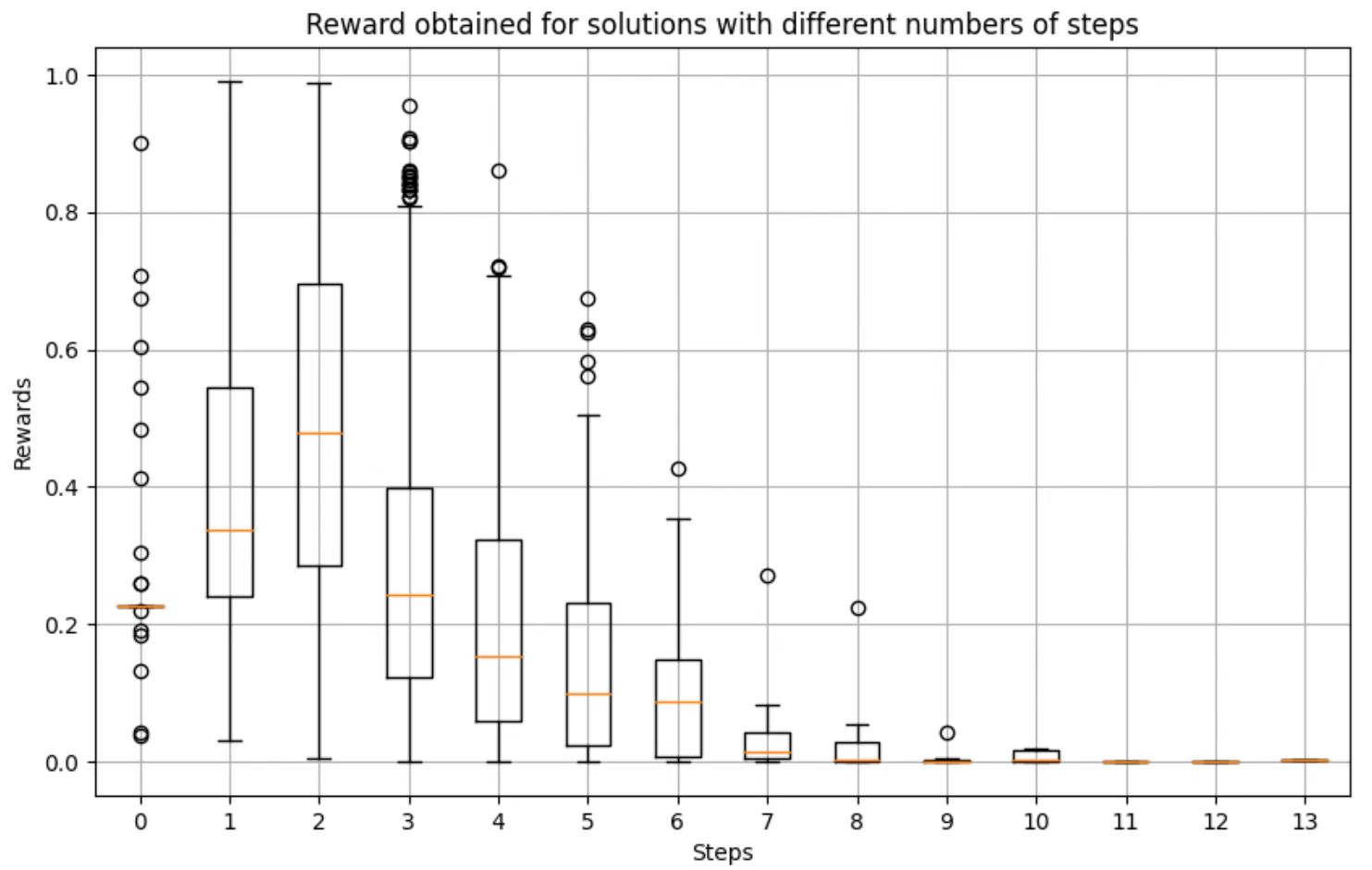}
    \caption{The relationship between the length of reasoning chains and rewards when nonlinearity is not incorporated into the reward scores of the reasoning process.}
    \label{fig:general_steps}
\end{minipage}
\begin{minipage}[t]{\linewidth}
    \centering
    \includegraphics[width=0.95\textwidth]{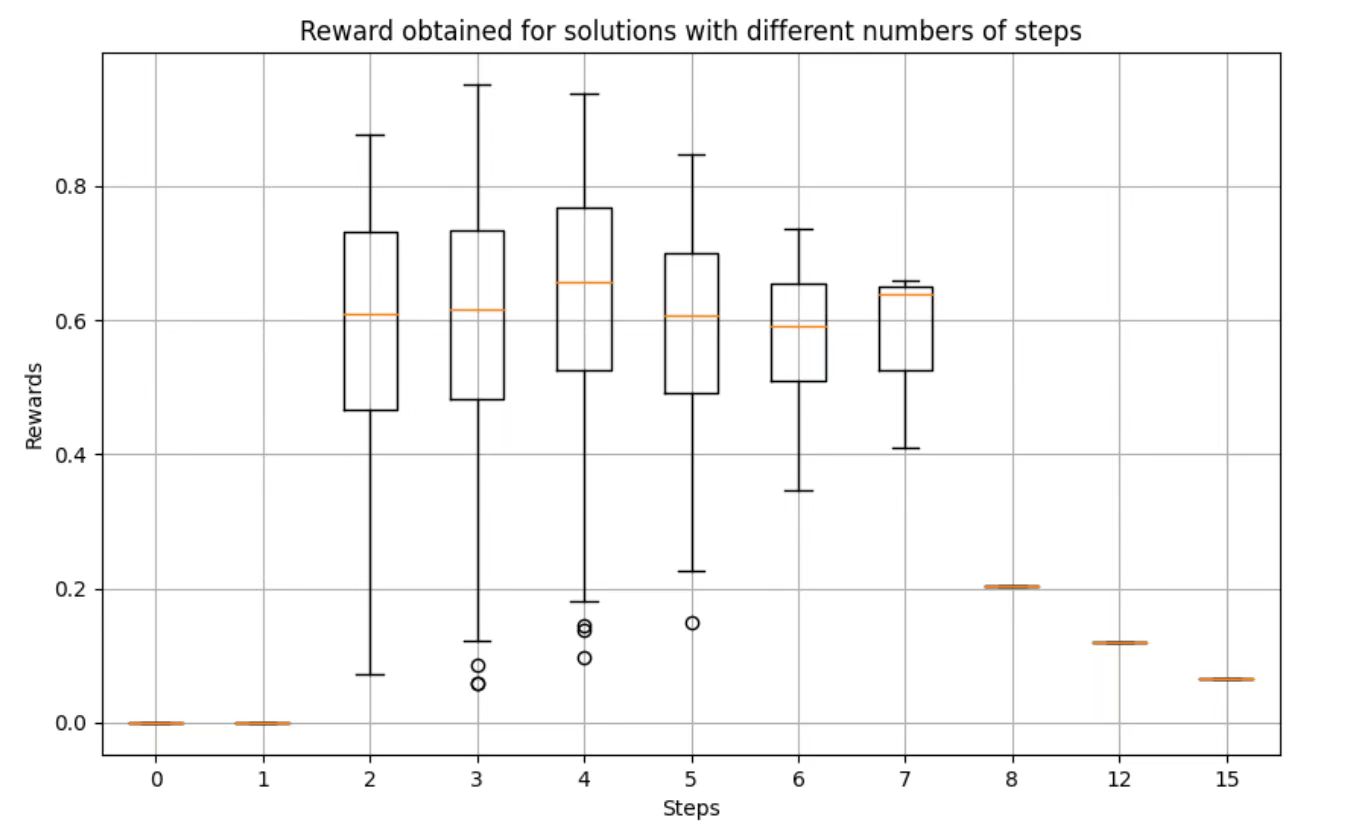}
    \caption{The relationship between the length of reasoning chains and rewards when nonlinearity is incorporated into the reward scores of the reasoning process.}
    \label{fig:nonlinear_steps}
\end{minipage}
\end{figure}

\paragraph{It is necessary to incorporate the length of reasoning chains into process supervision through nonlinear rewards.} Our ablation study confirms that process supervision depends not only on the accuracy of the reasoning chain but also on its length, and introduced nonlinear rewards accordingly. Specifically, in our experimental setup, we measure the length of reasoning chains by the number of reasoning steps involved. Table~\ref{tab:ablation} demonstrates that PSPO-WRS generates longer average reasoning chains. Further analysis of Figure~\ref{fig:general_steps} and Figure~\ref{fig:nonlinear_steps} reveals that without nonlinear rewards, the probability of the policy generating a particular reasoning process significantly decreases as the number of steps increases. Additionally, the reward scores for higher reasoning steps also diminish. However, the incorporation of nonlinear rewards mitigates this phenomenon.

Figure~\ref{fig:general_steps} demonstrates that without nonlinear rewards, the average reward score for reasoning chains declines once the number of steps exceeds three. This decline suggests that overlooking step count in process supervision can reduce overall reward scores, even when each step is accurately executed, due to simple multiplicative effects. Consequently, models may favor generating shorter reasoning processes.

Conversely, as shown in Figure~\ref{fig:nonlinear_steps}, after introducing nonlinear rewards, although the model still tends to generate multiple three-step reasoning processes, the proportion of reasoning processes with more steps has significantly increased. This phenomenon aligns with the prior knowledge incorporated during the reward shaping process. 
Furthermore, the model consistently yields high-scoring reasoning processes across different step counts, demonstrating its adaptability to tasks with variable reasoning lengths.

\section{Conclusion}

In this paper, we substantiate the critical role of accuracy and the length of reasoning chains in enhancing the efficacy of process supervision. We demonstrate through empirical evidence that these factors are interrelated in a nonlinear manner, significantly impacting the reward scores of reasoning processes. Inspired by these insights, we propose a novel process supervision paradigm, PSPO*, which systematically outlines the workflow from reward model training to policy optimization, and highlights the importance of nonlinear rewards in process supervision. To enhance the nonlinear impact, we propose using nonlinear accumulation function related to the length of reasoning chains, along with nonlinear reward shaping within the PSPO* paradigm. Based on the PSPO* paradigm, we introduced PSPO-WRS, which leverages an adjusted Weibull distribution for nonlinear reward shaping. The experimental results confirm our hypothesis and demonstrate that our method enables the model to generate more accurate and logically structured reasoning chains. Furthermore, these findings confirm the potential applicability of PSPO* paradigm across various reasoning disciplines.

\bibliography{aaai25}

\appendix

\section{Data Appendix}
\paragraph{Human-data Collection for Training Reward Model}

Training a robust reward model requires a balanced label distribution. While the steps generated by GPT-3.5 predominantly feature positive labels, we included additional reasoning step candidates from other LLMs, such as Abel-7b, to provide more negative examples and achieve label balance.  Human labelers would evaluate the given steps by their correctness, and correct answers to the question are provided as a reference.

\paragraph{Step Labelling Criteria} 

Each reasoning step is evaluated and assigned a label based on its correctness: `positive' (score of `1'), `neutral' (score of `0'), and `negative' (score of `-1'). A step receives a positive score if it accurately meets logical and computational requirements, correctly interprets the task, and contributes to deriving the correct answer. A neutral score is awarded if the step is correct but does not aid in reaching the correct conclusion. Conversely, steps that contain logical, computational, or factual inaccuracies, or are irrelevant to the given context and question, are assigned a negative score of -1.

\paragraph{Datasets} 

We adopt the MATH dataset, which includes AwpNLI, NewsNLI, RedditNLI, RTE-Quant, StressTest, and QQA datasets as reported by~\citet{chen-etal-2023-improving}. These datasets are further expanded using the GPT-3.5 API. The training dataset for the reward model is primarily composed of data labeled as `1'. To ensure a balanced dataset, steps labeled `0' and `-1' are replicated 2-3 times, yielding a final count of 16,587 positive, 11,072 neutral, and 16,236 negative steps. For evaluation, 20\% of the dataset is designated as test sets.

\end{document}